\documentclass[journal,twoside,web]{ieeecolor}
\makeatletter
\@ifpackageloaded{xcolor}{}{
    \usepackage{xcolor}
}
\makeatother

\usepackage{jsen}
\usepackage{cite}
\usepackage{amsmath,amssymb,amsfonts}
\usepackage{graphicx}
\usepackage{textcomp}
\usepackage{wrapfig}
\usepackage{multirow}
\usepackage{threeparttable}
\usepackage{graphicx}
\usepackage{algorithm}
\usepackage{algpseudocode}
\usepackage{subfigure}
\usepackage{textcomp,booktabs}
\usepackage{array}
\usepackage{calc}
\usepackage{hhline,tabu}
\usepackage{amsmath}
\usepackage{setspace}
\usepackage{siunitx}
\usepackage{breqn}
\usepackage{bm}
\usepackage{booktabs}
\usepackage{lineno,hyperref}
\usepackage{soul}

\soulregister\cite7 % 针对\cite命令
\soulregister\citep7 % 针对\citep命令
\soulregister\citet7 % 针对\citet命令
\soulregister\ref7 % 针对\ref命令
\soulregister\pageref7 % 针对\pageref命令

\def\BibTeX{{\rm B\kern-.05em{\sc i\kern-.025em b}\kern-.08em
    T\kern-.1667em\lower.7ex\hbox{E}\kern-.125emX}}
\definecolor{abstractbg}{rgb}{0.89804,0.94510,0.83137}
\begin{document}
\title{Cross-Modal Visuo-Tactile Representation Learning with Action Chunking Transformers for Contact-Rich Manipulation}
\author{Yaohua Liu, Rong Fu, Amir H. Gandomi, Simon Fong and Hengjun Zhang
\thanks{Manuscript received xx, xx, xx. (Corresponding author: Hengjun Zhang. E-mail: zhanghengjun@guet.edu.cn)}
\thanks{Yaohua Liu is with the Chinese Medicine Guangdong Laboratory, Hengqin, Zhuhai 519031, China (e-mail: liuyaohua@hqcmlab.cn).}
\thanks{Rong Fu is with the Institute of Collaborative Innovation, University of Macau, Macau 999078, China (e-mail: mc46603@um.edu.mo).}
\thanks{Amir H. Gandomi is with the Faculty of Engineering \& IT, University of Technology Sydney, Ultimo, NSW 2007, Australia, and also with the University Research and Innovation Center (EKIK), Óbuda University, 1034 Budapest, Hungary (e-mail: gandomi@uts.edu.au).}
\thanks{Simon Fong is with the Faculty of Science and Technology, University of Macau, Macau 999078, China (e-mail: ccfong@um.edu.mo). }
\thanks{Hengjun Zhang is with the School of Electronic Engineering and Automation, Guilin University of Electronic Technology, Guilin 541000, China.}
}

\IEEEtitleabstractindextext{%
\fcolorbox{abstractbg}{abstractbg}{%
\setlength{\intextsep}{0pt}%
\begin{minipage}{\textwidth}%
\begin{wrapfigure}[16]{r}{3in}%

\includegraphics[width=3in]{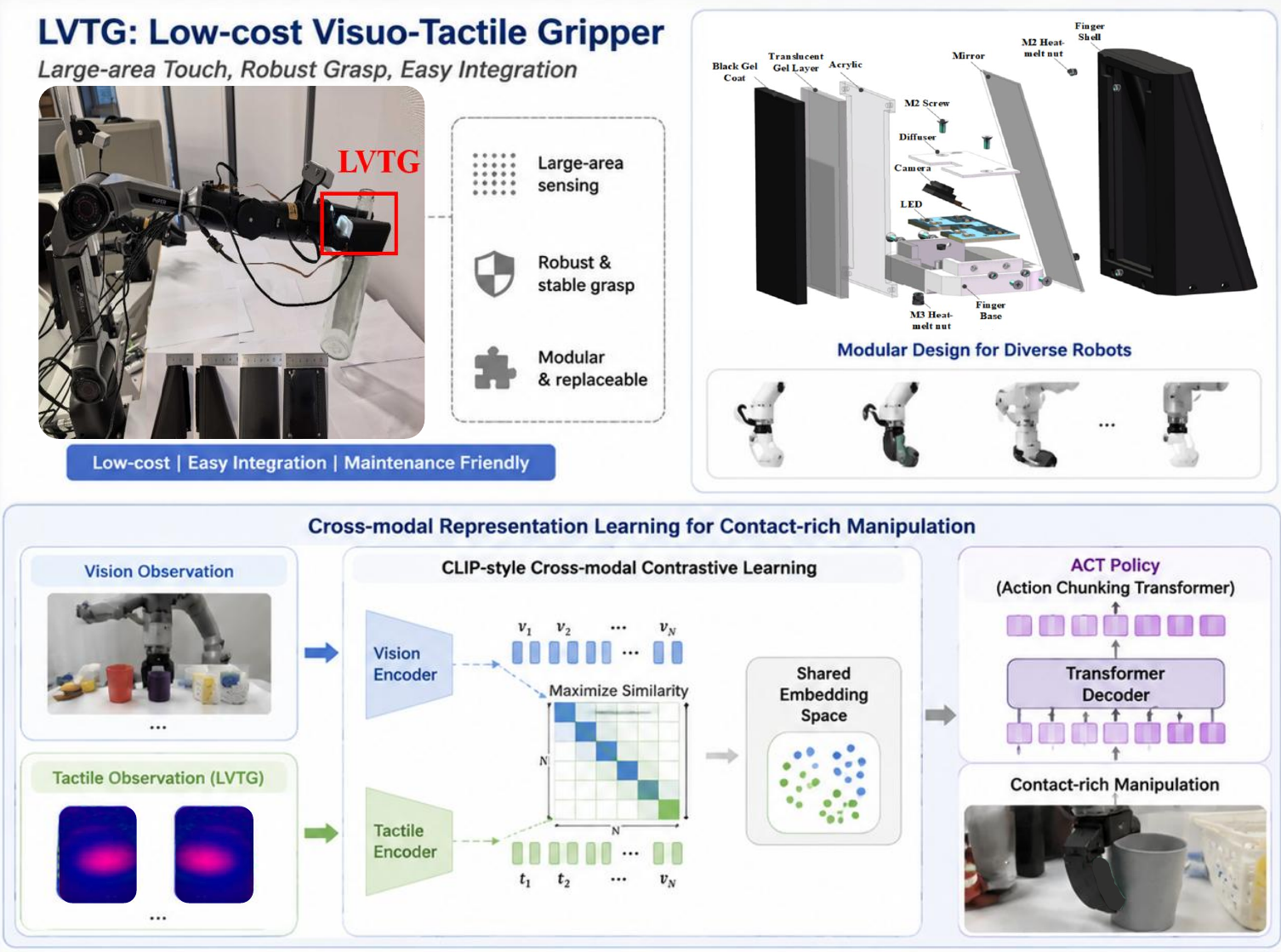}%
\end{wrapfigure}%

\begin{abstract}
Tactile feedback is important for contact-rich robotic manipulation, yet effective use of tactile observations remains challenging when tactile signals are image-like, hardware-dependent, and only weakly aligned with external visual observations. This study addresses this representation-learning problem by proposing a visuo-tactile contrastive learning framework for imitation-based manipulation. The method aligns external RGB observations and calibrated tactile images in a shared embedding space using a CLIP-style objective, and integrates the resulting representation into an Action Chunking Transformer (ACT) policy. A low-cost visuo-tactile gripper (LVTG) is proposed to provide a modular and durable sensing platform for reproducible data collection, supplying tactile observations that can be used by downstream manipulation algorithms. Experiments on contact-rich manipulation tasks show that tactile feedback improves the average task completion rate from 30$\%$ for a vision-only ACT baseline to 42$\%$, and that the proposed contrastive pretraining further increases the completion rate to 54$\%$. These results indicate that explicitly aligning visual and tactile observations provides more useful contact-aware features for downstream policy learning than directly adding tactile images without pretraining.
\end{abstract}

\begin{IEEEkeywords}
Action chunking transformer, contact-rich manipulation, cross-modal representation learning, robot learning, visuo-tactile sensing.
\end{IEEEkeywords}
\end{minipage}}}

\maketitle

\section{Introduction}
\label{sec:introduction}
\IEEEPARstart{H}{umans} possess a remarkable capacity to perform complex manipulation tasks that involve rich physical contact, such as handling fragile or deformable objects, by effectively fusing visuo and tactile information. Vision plays a key role in object recognition and spatial localization, while tactile sensing provides real-time feedback on local contact conditions, including surface texture, material compliance, and force distribution, which are typically inaccessible through vision alone \cite{yoshioka2007texture, tiest2009cues, ma2019dense}. For instance, when interacting with delicate items, visuo cues guide the initial reaching and grasping motions, and tactile signals enable continuous, fine-grained adjustments to grip strength and orientation to prevent damage. Replicating this coordinated use of multimodal sensory input in robotic systems remains a significant challenge, particularly for tasks that require high precision and rapid responsiveness to changing contact dynamics.

Recent advances in vision-based tactile sensors have significantly improved robotic performance in contact-rich manipulation by providing high-resolution feedback on physical interactions, such as object deformation and force distribution \cite{lin2022dtact, lin20239dtact, yuan2017gelsight}. Despite these improvements, their widespread deployment remains limited by practical constraints, including high manufacturing costs, bulky designs, and poor compatibility with diverse robotic platforms. In addition, many existing systems suffer from limited sensing coverage and narrow fields of view, which restrict their effectiveness in tasks requiring both dexterity and fine-grained perception \cite{wilson2023cable, li2024vision, pan2022algorithms}. These limitations are particularly critical in applications such as agricultural manipulation, where handling soft and fragile objects demands robust and reliable tactile perception \cite{niu2025deep,mandil2023tactile}. However, current visuotactile sensors often lack durability and modularity, making them unsuitable for long-term operation and rapid maintenance in real-world environments. Furthermore, their high cost hinders large-scale adoption, underscoring the need for an integrated, low-cost visuotactile gripper with large sensing coverage, robustness, and modular design.

While vision-based imitation learning (IL) has shown strong promise in robotic manipulation tasks \cite{chi2023diffusion, zhao2023learning, black2410pi0}, its effectiveness in contact-rich settings remains limited by the absence of tactile feedback during training. This omission impairs a robot's capacity to react promptly to environmental perturbations, particularly in scenarios demanding rapid sensory response and adaptive force modulation. Current tactile data collection systems have struggled to establish an effective link between visuo observations and tactile signals. Many rely either on motion capture or teleoperation setups that introduce substantial latency and require intricate calibration procedures \cite{wu2025freetacman, xue2025reactive}. Furthermore, handheld approaches that convey tactile information through mechanical linkages often suffer from degraded signal fidelity and slower response times, further limiting their utility in real-time manipulation.

In this work, we introduce a unified visuotactile embodied manipulation framework to address the aforementioned limitations and enable stable, efficient, and cost-effective manipulation in contact-rich environments. The central contribution is a cross-modal representation learning pipeline that transfers information between vision and touch and improves downstream policy learning. The visuo-tactile gripper serves as a practical data-collection and validation platform for this pipeline, providing paired observations for learning without requiring high-resolution tactile sensing. 

The contributions of the paper are summarised as:
\begin{enumerate}
  \item We propose a cross-modal representation learning framework that leverages visuo-tactile feedback to learn informative contact-enhanced representations without requiring high-resolution tactile sensing. A CLIP-style contrastive objective is adopted to align visual and tactile observations in a shared embedding space, facilitating effective cross-modal perception for manipulation.
  \item We integrate the learned representations with an Action Chunking Transformer (ACT) policy and show that the cross-modal features improve downstream contact-rich manipulation.
  \item We provide a low-cost visuo-tactile gripper implementation (LVTG) as a reproducible platform for data collection and evaluation, supporting the algorithmic contribution.
\end{enumerate}

\section{Related Works}
\subsection{Tactile Sensing for Robotic Manipulation}
Tactile sensing is essential for enabling robots to perform precise manipulation in contact-rich environments. Conventional approaches typically employ force/torque sensors mounted at the wrist or within robotic joints \cite{cao2021six, muscolo2023force, li2024multi}. Although these sensors deliver direct measurements of interaction forces, they are often costly, susceptible to noise, and incapable of resolving fine-grained local contact information. As alternatives, various electrical tactile sensors have been investigated, including those based on capacitive \cite{kim2011capacitive, maiolino2013flexible}, resistive \cite{zhu2022recent, shu2024flexible}, MEMS \cite{bayer2022mems}, and magnetic principles \cite{man2022recent, alfadhel2015magnetic}. While these technologies enable compact integration, they frequently face limitations such as low spatial resolution, sensitivity to environmental disturbances, and complex or non-scalable fabrication processes.

In contrast, optical tactile sensors such as GelSight, GelSlim, MC-Tac, and 9DTact \cite{dong2017improved, Ren2023MC-Tac, gelsight-mini, lin20239dtact} utilize embedded cameras to capture high-resolution images of elastomer deformation. These systems can recover rich contact information, including surface texture, local geometry, and force distribution, rendering them highly effective for contact-rich manipulation tasks. Nevertheless, many existing vision-based tactile sensors are implemented as monolithic, custom-fabricated units that lack modularity, making rapid replacement or adaptation to different robotic end-effectors impractical. Their relatively high cost and mechanical fragility also hinder broad deployment in real-world robotic applications, motivating tactile platforms that combine optical sensing quality with low-cost, maintainable hardware.

\subsection{Dataset Collection Systems for Robot Learning}
The success of robot learning, particularly imitation learning, critically relies on the availability of large-scale, high-quality demonstration datasets. To collect visuomotor data, various teleoperation approaches have been developed, including motion capture systems enhanced with AR/VR visualization \cite{xue2025reactive, chen2025arcap}, 3D space mice \cite{bolanos2021three, martin2021jrdb}, master-slave robotic platforms \cite{aldaco2024aloha, fu2024mobile}, and wearable interfaces \cite{wu2025freetacman, chi2024universal}. Although these systems can record accurate expert trajectories, they often require intricate calibration, incur high costs, or provide limited tactile feedback. Handheld collection devices improve flexibility across robot embodiments, but rod-based or trigger-actuated grippers can introduce mechanical backlash and indirect force transmission, reducing the fidelity of contact observations.

Recent systems have begun to integrate tactile sensing into data collection for contact-rich manipulation. Reactive diffusion policy \cite{xue2025reactive} combines AR-based teleoperation with slow--fast visual--tactile learning, showing that high-frequency tactile feedback can improve robustness and generalization. FreeTacMan \cite{wu2025freetacman} further couples tactile sensing with human operators for in-situ demonstration collection. Vision-based tactile grippers can provide richer contact images than force-only interfaces \cite{abad2020visuotactile}, but practical dataset collection still requires hardware that is inexpensive, durable, easy to repair, and adaptable to varied objects, tasks, and robot platforms.

To support robot-learning data collection, our LVTG gripper is designed as a low-cost, modular, and easily fabricable vision-based tactile sensing platform. Its enlarged sensing area and replaceable skin module improve practical durability and maintenance, while its compact structure allows integration into downstream manipulation pipelines. By providing paired visual and tactile observations during teleoperated manipulation, the platform lowers the barrier to collecting reproducible visuo-tactile datasets for learning contact-aware policies.

\subsection{Visuo-Tactile Manipulation Learning Algorithms}
Vision-based imitation learning has achieved strong results in manipulation through action-chunking or diffusion-based policies \cite{zhao2023learning, chi2023diffusion}, but contact-rich tasks often require tactile feedback to resolve local contact conditions that are occluded or visually ambiguous. Recent visuo-tactile algorithms therefore incorporate touch either as an additional policy input or as a supervisory signal for representation learning. VITaL pretraining \cite{george2025vital} shows that visuo-tactile pretraining can improve both tactile and non-tactile manipulation policies, suggesting that tactile observations can provide useful learning signals even when touch is not available at inference. Reactive Diffusion Policy \cite{xue2025reactive} further combines a slow visual policy with a fast tactile feedback pathway, allowing contact feedback to correct actions during execution. T-Dex \cite{guzey2023dexterity} uses self-supervised tactile representations learned from robotic play and combines them with visual observations for dexterous manipulation, highlighting the value of pretraining high-dimensional tactile signals before policy learning.

Beyond policy architectures, several studies focus on learning reusable tactile or visuo-tactile representations. Self-Supervised Visuo-Tactile Pretraining \cite{kerr2022self} aligns spatially paired visual and tactile observations through cross-modal contrastive learning for deformable-object tasks. Sparsh \cite{higuera2024sparsh} learns general-purpose touch representations from large-scale vision-based tactile data, showing that self-supervised tactile pretraining can outperform task-specific training across multiple tactile perception and manipulation benchmarks. Robot Synesthesia \cite{yuan2023robot} represents tactile feedback as point-cloud information and fuses it with vision for in-hand manipulation, while ConViTac \cite{wu2025convitac} uses contrastive embeddings to condition visual--tactile feature fusion through cross-modal attention. These methods demonstrate the importance of representation learning for using tactile images effectively, but many depend on specialized sensors, large pretraining corpora, or task-specific fusion modules. In contrast, this work adopts a compact CLIP-style alignment objective using paired visual and tactile observations collected from a low-cost gripper, and evaluates whether the resulting contact-aware representation improves an ACT policy in contact-rich manipulation.

\section{Materials and Methods}
\subsection{Hardware Design and Fabrication}
\begin{figure}[h]
    \centering
    \includegraphics[width=.9\linewidth]{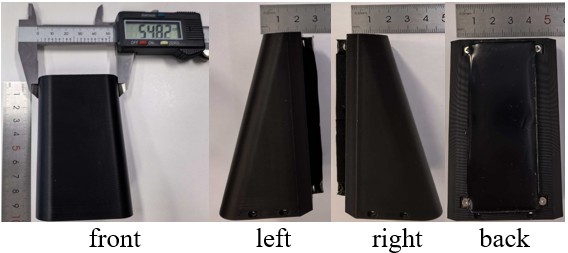}
    \caption{The figure of LVTG's form factor.}
    \label{Fig: size}
\end{figure}

The hardware is treated here as an enabling sensing front-end for the learning pipeline. Its role is to provide repeatable paired visual and tactile observations, while keeping the geometry simple enough for consistent calibration and data collection. As illustrated in Figure~\ref{Fig: size}, LVTG features a rectangular form factor with an integrated sloped surface, which enhances its adaptability to various object shapes. The gripper measures approximately 54.9 mm in width and 90.6 mm in length. The end-effector employs a parallel four-bar linkage mechanism to realize a two-finger parallel-jaw gripper. The linkage is actuated by a motor located at the robot wrist, which pulls the four-bar structure to drive the symmetric opening and closing of the two fingers.

\begin{figure*}[!th]
    \centering
    \includegraphics[width=.8\linewidth]{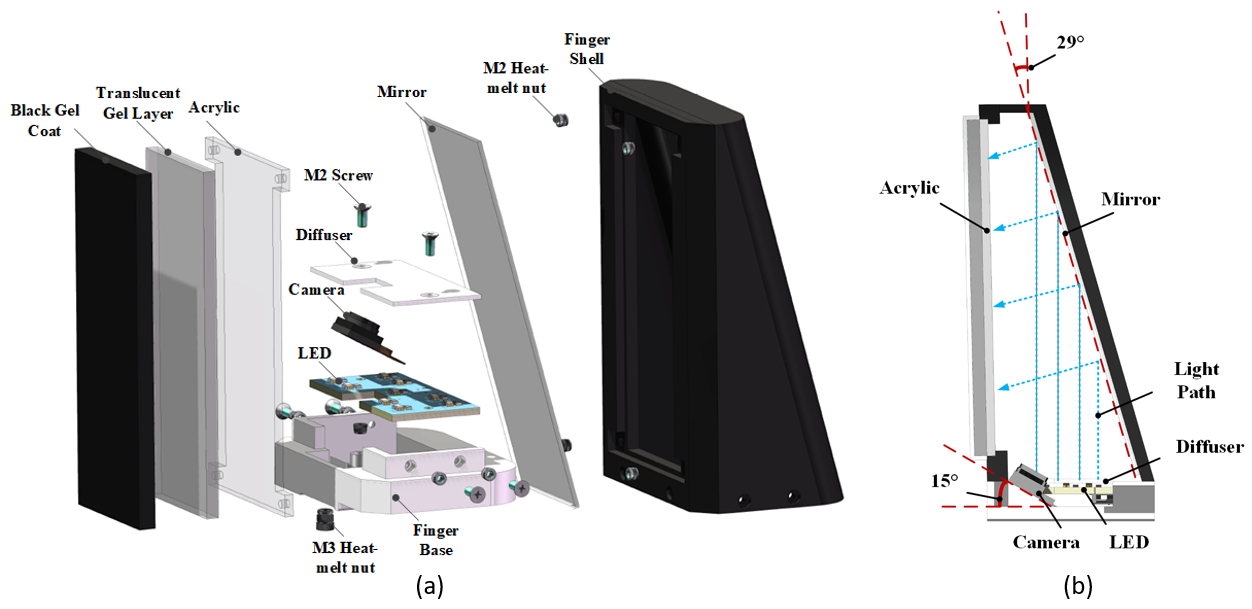}
\caption{(a) Exploded mechanical view of LVTG. (b) Illustration of the LVTG optical path.}
    \label{Fig: olvtg}
\end{figure*}

As shown in Figure~\ref{Fig: olvtg}(a), each visuo-tactile finger comprises a replaceable skin module, an LED illumination board, a camera module, a diffuser, a mirror, a finger shell, and a finger base. The design follows three practical constraints that are important for computational evaluation: the optical path must cover the tactile surface, the skin must be replaceable without disassembling the optical components, and the fabrication route must rely on accessible materials and standard manufacturing processes. Table~\ref{tab:bom} lists the main materials and fabrication processes. Based on retail component prices, the material cost of one tactile finger is kept below 12 USD. This estimate covers the listed sensor components and consumables, but excludes the robot arm, gripper actuator, labour, shipping, and general-purpose fabrication tools.

\begin{table}[tbp]
\centering
\caption{Bill of materials (BOM) and fabrication processes for one LVTG finger}
\resizebox{\linewidth}{!}{ 
\begin{tabular}{ccc}
\toprule
Component       & Description             & Process                                                                        \\ \hline
Black Gel       & Smooth-On Psycho Paint  & \multirow{2}{*}{\begin{tabular}[c]{@{}c@{}}Silicone\\ processing\end{tabular}} \\
Translucent Gel & Smooth-On Ecoflex 00-10 &                                                                                \\ \hline
Acrylic         & 2mm thick acrylic board & Laser cutting                                                                  \\ \hline
Mirror          & 30x65x2mm               & \multirow{2}{*}{PCB soldering}                                                 \\
LED Board       & 42x21x1.5mm             &                                                                                \\ \hline
Screws          & 10 M2-6                 & \multirow{4}{*}{Off-the-shelf}                                                 \\
Nets            & 8 M2-3-5                &                                                                                \\
Glue            & AD-119                  &                                                                                \\
Camera          & OV5674                  &                                                                                \\ \bottomrule
\end{tabular}
}
\label{tab:bom}
\end{table}

\subsubsection{Skin Module}
The tactile skin follows the optical deformation principle of 9DTact \cite{lin20239dtact}, but is redesigned for larger contact coverage, stronger layer adhesion, and module-level replacement. It consists of a black gel coat, a translucent gel layer, and an acrylic substrate (Figure~\ref{Fig: olvtg}). The translucent silicone gel (Smooth-On Ecoflex 00-10) is moulded directly onto a 2 mm acrylic substrate treated with DOWSIL 1200 OS Primer, which improves adhesion between the compliant layer and the rigid support. The black surface layer is prepared using Smooth-On Psycho Paint, silicone pigment, and Novocel Matte at a 4:4:1:2 ratio, then cast onto the translucent gel surface. A 3 mm gel thickness is selected as a design compromise: thinner layers increase sensitivity but are more vulnerable to tearing and saturation, whereas thicker layers improve durability but attenuate small contact deformations. The resulting skin module can be produced in batches and replaced independently of the camera, LED board, and finger housing.

\subsubsection{LED Module}
The illumination module is powered through a standard 5 V Micro-USB interface. It contains nine high-intensity LEDs (TZ-P2-1206WYCS2-0.9T; luminous intensity 800--9000 mcd; colour temperature 10,000--15,000 K), each stabilised with a 10 $\Omega$ resistor. A constant-current driver (CN5711) reduces brightness drift across LEDs. This arrangement limits the influence of ambient light and provides repeatable image acquisition for the learning pipeline.

\subsubsection{Camera Module and Optical Path}
As shown in Figure~\ref{Fig: olvtg}(b), a 3D-printed diffuser first homogenises the LED illumination. A 30 mm $\times$ 65 mm mirror then folds the optical path, reflecting light towards the tactile skin and projecting contact-induced deformation back to the camera. The OV5674 camera has a 120$^\circ$ field of view and is mounted at a 15$^\circ$ angle relative to the finger base plane. The folded optical path therefore supports tactile imaging without increasing the finger thickness to the level of many conventional camera-in-fingertip designs.

\subsubsection{Finger Housing and Assembly}
The finger shell and base are fabricated by 3D printing. The shell is printed in black material to reduce stray light, while the base provides the mechanical mounting platform for the camera, LED board, and diffuser. Assembly proceeds as follows. First, the mirror is fixed in the inner shell slot to define the optical path. Second, the skin module is attached to the shell using four M2 screws and heat-set nuts, allowing replacement after wear without disturbing the camera calibration. Third, the LED board and diffuser are mounted on the finger base, and the camera is aligned at the 15$^\circ$ angle before being fixed in place. Finally, the shell and base are fastened with four M2 screws.

\subsection{Tactile Image Calibration and Representation}
\begin{figure}[h]
    \centering
    \includegraphics[width=.85\linewidth]{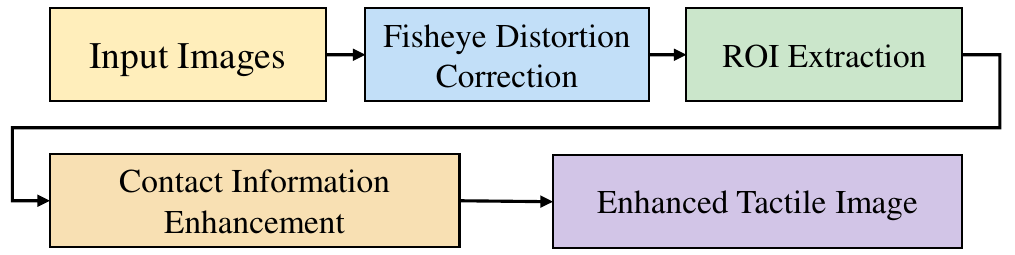}
\caption{Overview of the tactile image calibration and enhancement pipeline.}
    \label{Fig: pp}
\end{figure}

To make the hardware output usable for learning, raw tactile images are converted into a calibrated contact representation before training. The processing pipeline contains three reproducible stages, as shown in Figure~\ref{Fig: pp}: fisheye camera distortion correction, region of interest (ROI) extraction, and contact information enhancement. This calibration stage is part of the algorithmic pipeline because it defines the representation consumed by the contrastive learner.

\begin{table*}[!t]
\centering
\caption{Comparison of different vision-based tactile sensors. (*Manufacturing of 1000 pieces. $\dagger$ Commodity price.)}

\begin{tabular}{l c c c c}
\toprule
Sensor & Sensing Area [$mm^2$] $\uparrow$ & Weight [$g$] $\downarrow$ & FPS $\uparrow$ & Cost[$\$ $] $\downarrow$ \\
\midrule
GelSight \cite{dong2017improved}  & 252 &  NA & \textbf{90} & 30 \\
GelSlim 3.0 \cite{taylor2022gelslim}  & 675 &  45 & \textbf{90} & 25* \\
DIGIT \cite{lambeta2020digit}  & $19 \times 16 = 304$ &  \textbf{20} & 60 & 15* / 300$\dagger$ \\
GelSight-Mini \cite{gelsight-mini}  & $19 \times 15 = 285$ & 20.8 & 25 & 499$\dagger$ \\
DTact \cite{lin2022dtact} & $24 \times 24 = 576$ &  78 & 60 & 34 \\
9DTact \cite{lin20239dtact} & $24 \times 18 = 432$ & \textbf{20} & \textbf{90} & 15 \\
\midrule
\textbf{LVTG (Ours)}& $80 \times 30 = \textbf{2400}$ & 70 & \textbf{90}& \textbf{12}\\
\bottomrule
\end{tabular}
\label{tab:sensors}
\end{table*}

\subsubsection{Fisheye Camera Distortion Correction.} The employed camera module (OV5674) has a wide field of view (120$^\circ$), which introduces severe distortion, especially near the image boundaries. To prevent distortion from degrading the quality of the captured contact signals, we perform fisheye undistortion before deploying the camera into the sensor housing. Specifically, we capture 20 images of a 7$\times$10 checkerboard pattern with a 15 mm square size and use OpenCV's fisheye calibration routine to estimate the camera intrinsic matrix $K$ and distortion coefficients $D=(k_1,k_2,k_3,k_4)$. In the fisheye model, a point is first mapped using
\begin{equation}
\theta_d=\theta(1+k_1\theta^2+k_2\theta^4+k_3\theta^6+k_4\theta^8),
\quad
\begin{bmatrix}
u\\v\\1
\end{bmatrix}
\sim
K
\begin{bmatrix}
x\theta_d/r\\y\theta_d/r\\1
\end{bmatrix},
\end{equation}
where $(x,y)$ denotes the normalised image coordinate, $r=\sqrt{x^2+y^2}$, and $\theta=\arctan(r)$. The calibrated $K$ and $D$ are then used to generate a remapping table, which is fixed for all subsequent tactile image acquisition.

\subsubsection{ROI Extraction.} Since only the gel surface contains meaningful tactile information, it is necessary to exclude irrelevant regions from the raw images. We achieve this through a three-step process: (i) a mask $M(x,y)$ of the gel surface region is generated during a one-time calibration step by selecting the four gel-surface corner points; (ii) the selected points are saved and reused for subsequent recordings; (iii) an affine transformation and pixel slicing are applied to rectify the gel region into a rectangular patch, which can be expressed as,
\begin{equation}
\begin{bmatrix}
x' \\ y' \\ 1
\end{bmatrix}
=
A
\begin{bmatrix}
x \\ y \\ 1
\end{bmatrix},
\quad
A \in \mathbb{R}^{2\times 3},
\end{equation}
where $(x,y)$ and $(x',y')$ denote original and rectified coordinates, respectively. This process yields a clean ROI corresponding precisely to the tactile surface, minimising interference from the external background while avoiding repeated manual adjustment during data collection.

\subsubsection{Contact Information Enhancement.} Due to the bottom-mounted LED module, vertical brightness imbalance is naturally introduced across the tactile images. To correct illumination variation and emphasise contact-induced features, we implement a contact information enhancement procedure. First, a vertical attenuation mask $W(y)$ is applied to the ROI image to normalise brightness gradients, which can
be described as, 
\begin{equation}
I'(x,y) = I(x,y) \cdot W(y), \quad
W(y) = \exp\!\left(-\alpha \frac{y}{H}\right),
\end{equation}
where $H$ is the ROI height and $\alpha$ controls the decay rate. In practice, $\alpha$ is selected once from the no-contact reference frame to reduce the vertical brightness gradient and is then kept fixed. Given the corrected reference image $I'_{\text{ref}}$ and current tactile image $I'_{\text{cur}}$, we compute,
\begin{equation}
I_{\text{dark}} = \max(0, I'_{\text{ref}} - I'_{\text{cur}}), 
I_{\text{bright}} = \max(0, I'_{\text{cur}} - I'_{\text{ref}}).
\end{equation}

Finally, similar to the approach in 9DTact, the tactile ROI is compared with a pre-recorded reference image to compute a darker channel and a brighter channel, highlighting local deformations. The darker channel, brighter channel, and corrected reference channel are combined into a composite enhanced tactile image, which can be described as,
\begin{equation}
I_{\text{enh}} = \text{Concat}(I_{\text{dark}}, I_{\text{bright}}, I'_{\text{ref}}),
\end{equation}
where $\text{Concat}(\cdot)$ denotes channel-wise concatenation, and it can serve as the final input for downstream learning algorithms.

\section{Cross-Modal Representation and Policy Learning}
\subsection{Learning Formulation}
\begin{figure*}[!t]
    \centering
    \includegraphics[width=.7\linewidth]{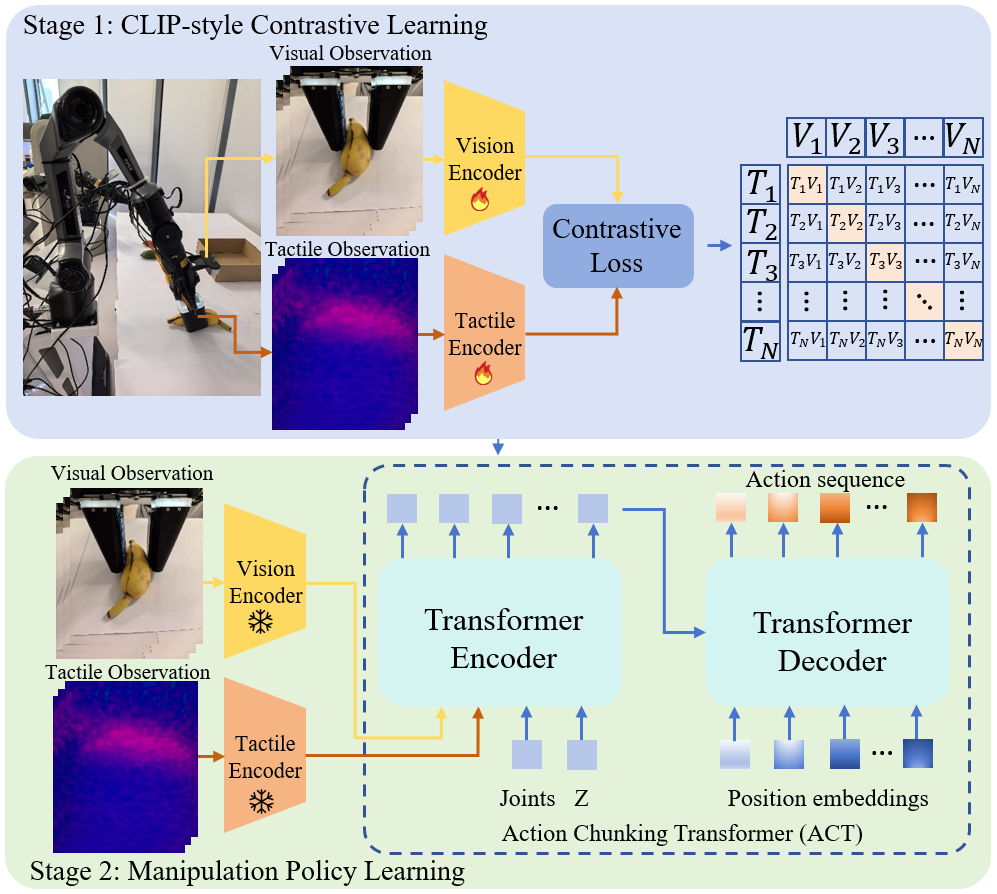}
\caption{Overview of the manipulation learning pipeline.}
    \label{Fig: act}
\end{figure*}

We formulate manipulation learning as a two-stage computational framework, comprising visuo-tactile contrastive pretraining followed by action-chunking policy optimisation, as illustrated in Figure~\ref{Fig: act}. Each training sample is denoted as $(I_i, T_i, q_i, A_i)$, where $I_i$ is the external RGB observation, $T_i$ is the enhanced tactile image generated by the calibration pipeline, $q_i$ is the normalised 7-DOF robot joint state, and $A_i=\{a_i,\ldots,a_{i+H-1}\}$ is an action chunk of horizon $H$. Although $T_i$ is represented as an image, it is generated from contact-induced deformation rather than optical appearance; we therefore treat it as an image-like tactile representation and learn a shared visuo-tactile embedding with the external visual stream. This formulation makes the tactile representation explicitly conditioned on the robot state, which is important because the same contact pattern can imply different manipulation states under different finger poses.

Although tactile signals are image-like, directly applying RGB-pretrained vision encoders can produce weak contact features because tactile images differ from natural images in texture, illumination statistics, and semantic content \cite{george2025vital}. We therefore adopt a CLIP-style contrastive objective \cite{radford2021learning} to align external visual observations and tactile observations acquired at the same timestamp. The objective is not intended to reconstruct tactile deformation from RGB images; instead, it encourages the two modalities to encode shared manipulation states, such as object pose, grasp phase, and contact occurrence. The visual encoder $f_v$ and tactile encoder $f_t$ use ResNet backbones initialised from the same checkpoint. Projection heads $g_v$ and $g_t$ map the encoded features to a shared embedding space. The tactile branch concatenates tactile features with the normalised joint vector before projection:
\begin{equation}
v_i=\frac{g_v(f_v(I_i))}{\lVert g_v(f_v(I_i))\rVert_2}, \quad
t_i=\frac{g_t([f_t(T_i);q_i])}{\lVert g_t([f_t(T_i);q_i])\rVert_2}.
\end{equation}
For a mini-batch of $B$ paired samples, $(v_i,t_i)$ is treated as the positive pair, while unmatched cross-modal pairs in the same mini-batch serve as negatives. The symmetric contrastive loss is defined as
\begin{equation}
\small
\begin{aligned}
\mathcal{L}_{\text{con}}
=-\frac{1}{2B}\sum_{i=1}^{B}
\Bigg(
&\log\frac{\exp(v_i^\top t_i/\tau)}
{\sum_{k=1}^{B}\exp(v_i^\top t_k/\tau)} \\
&+\log\frac{\exp(t_i^\top v_i/\tau)}
{\sum_{k=1}^{B}\exp(t_i^\top v_k/\tau)}
\Bigg),
\end{aligned}
\label{eq:contrastive_loss}
\end{equation}
where $\tau$ is the learnable temperature parameter. Mild geometric and photometric augmentations, including scaling, random cropping, and brightness perturbation, are applied consistently during pretraining to improve robustness to viewpoint variation, local deformation, and sensor noise. Flipping operations are avoided when they would invert task-relevant spatial semantics.

In the policy learning stage, the pretrained encoders extract visual and tactile features that are concatenated with proprioceptive information and fed into an ACT \cite{zhao2023learning}. The same encoder architecture and policy backbone are used for all ablation settings so that performance differences mainly reflect the availability of tactile input and contrastive pretraining. The policy predicts a sequence of absolute joint commands rather than a single next-step action:
\begin{equation}
\hat{A}_i=\pi([v_i;t_i;q_i]), \quad
\mathcal{L}_{\text{policy}}=\frac{1}{H}\sum_{h=0}^{H-1}\lVert a_{i+h}-\hat{a}_{i+h}\rVert_1+\beta\mathcal{L}_{\text{KL}},
\end{equation}
where $\mathcal{L}_{\text{KL}}$ is the latent regularisation term used in ACT when the conditional variational branch is enabled. This two-stage design separates contact representation learning from policy optimisation, which improves sample efficiency and makes it possible to evaluate the contribution of tactile pretraining independently from the downstream controller.

\subsection{Cross-Modal Contrastive Learning}
As shown in algorithm~\ref{alg:lvtg}, the central algorithmic contribution is the cross-modal alignment of tactile and visual observations. The goal is to learn a representation in which paired visual and tactile frames are close in embedding space, while mismatched pairs are separated. This makes contact state available to the policy in a form that is easier to generalise than raw tactile pixels.

We use paired observations from the same time step as positive samples and in-batch mismatched pairs as negatives. The visual encoder processes the external RGB frame, while the tactile encoder processes the calibrated tactile image together with the robot state. Conditioning the tactile branch on proprioception helps disambiguate identical contact patterns that occur under different grasp poses.

The training objective is intentionally simple: a symmetric contrastive loss aligns the two modalities, and the learned encoders are then frozen for downstream policy learning. This separation isolates the representation-learning effect from policy optimisation and makes the contribution of tactile pretraining easier to measure. In practice, the learned embeddings act as compact contact-state descriptors that are more informative than either raw RGB or tactile input alone.

% \subsection{Policy Adaptation}
The second stage maps the learned multimodal representation to action chunks with ACT. Compared with end-to-end training from scratch, this staged design reduces the burden on the policy network because it receives a contact-aware latent code rather than raw multimodal inputs. The policy therefore focuses on action synthesis, while the contrastive encoder handles multimodal alignment.

% This design also makes the ablation study more interpretable. The vision-only baseline tests whether external RGB observations are sufficient. The tactile baseline tests whether adding tactile input alone is enough. The pretrained model tests whether the learned cross-modal representation provides an additional gain beyond raw tactile sensing.

\begin{algorithm}[t]
\caption{Two-Stage Visuo-Tactile Manipulation Learning}
\label{alg:lvtg}
\begin{algorithmic}[1]

\State \textbf{Input:} Dataset $\mathcal{D}=\{(I_i, T_i, q_i, A_i)\}$, temperature $\tau$, chunk horizon $H$
\State \textbf{Output:} Visual encoder $f_v$, Tactile encoder $f_t$, policy $\pi$

\State \textbf{Stage 1: Contrastive Learning}
\State Initialize visual encoder $f_v$, tactile encoder $f_t$
\State Initialize projection heads $g_v, g_t$

\For{each iteration}
    \State Sample mini-batch $\{(I_i, T_i, q_i)\}_{i=1}^{B}$
    \State $v_i \gets \text{norm}(g_v(f_v(I_i)))$
    \State $t_i \gets \text{norm}(g_t([f_t(T_i);q_i]))$
    \State Define matched pairs $(v_i, t_i)$ as positives
    \State Use unmatched in-batch pairs as negatives
    \State Compute $\mathcal{L}_{\text{con}}$ using Eq.~\eqref{eq:contrastive_loss}
    \State Update $f_v, f_t, g_v, g_t$
\EndFor

\State \textbf{Stage 2: Policy Learning}
\State Freeze visual encoder $f_v$ and tactile encoder $f_t$

\For{each iteration}
    \State Sample $(I_i, T_i, q_i, A_i)$
    \State Extract features $v_i, t_i$
    \State $z_i \gets \text{concat}(v_i, t_i, q_i)$
    \State $\hat{A}_i \gets \pi(z_i)$
    \State Update $\pi$ using $\mathcal{L}_{\text{policy}}$
\EndFor

\State \Return $\pi$

\end{algorithmic}
\end{algorithm}

\section{Experiment results and discussion} \label{Sec: experiment}
The evaluation is organised around two questions that separate system-level sensing reliability from policy learning performance. First, we assess the grasp stability and reliability of LVTG under teleoperated manipulation and compare it with representative vision-based tactile sensor configurations. Second, we examine whether cross-modal visuo-tactile pretraining improves ACT policy learning in contact-rich manipulation tasks.

\subsection{Grasp Stability and Reliability}
We first evaluate the grasp stability and operational reliability of LVTG in teleoperated manipulation. This experiment does not involve an autonomous grasping policy or learning-based grasp planner. Instead, it measures whether different vision-based tactile sensor configurations can support stable manual grasping under a consistent teleoperation protocol. The experimental setup includes a set of 20 daily objects with varying shapes, sizes, weights, and surface properties (e.g., cylindrical bottles, boxes, deformable sponges, and fragile items such as eggshell containers). Each object is grasped 30 times under identical conditions, and the success rate is defined as the percentage of trials in which the object is lifted and held stably for at least 10 seconds without slipping. To ensure a consistent comparison across sensor configurations, grasp poses are specified using the same gripper-agnostic protocol. Specifically, for each object, we predefine a set of feasible grasp configurations based on its geometry and accessible approach directions, without sensor-specific tuning. These predefined grasp poses are then applied identically to all tested configurations.
\begin{figure*}[!t]
    \centering
    \includegraphics[width=.8\linewidth]{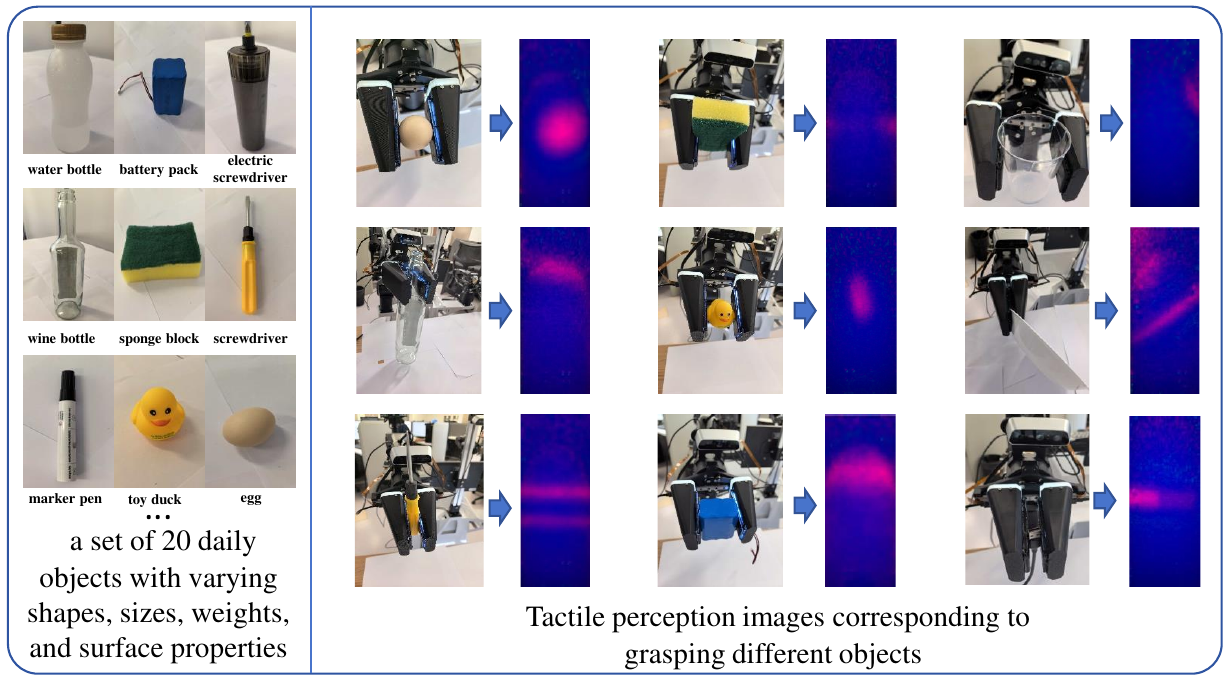}
    \caption{Tactile perception results during grasping different objects.}
    \label{Fig: tac_pic}
\end{figure*}

As shown in Fig. \ref{Fig: tac_pic}, LVTG can stably grasp objects of different shapes and materials, such as fragile eggs and transparent wine bottles, and can clearly perceive the tactile conditions at the contact points between the gripper and the objects. This provides tactile feedback for embodied manipulation and data collection. Due to its larger gripping contact area and greater opening angle, LVTG can more stably grasp unstructured objects as well as heavy and slippery objects. In particular, objects with smooth or deformable surfaces benefit from the extended tactile sensing range, which helps maintain contact coverage during teleoperated grasping and provides more informative tactile images for subsequent learning. It is noted that the two fingers in the LVTG are symmetrically designed, and their sensing modules share identical structure and configuration. Therefore, the tactile images captured from both fingers are visually very similar under comparable contact conditions.

To further evaluate the performance of LVTG relative to other visuo-tactile sensors in terms of grasp stability and generalizability, we compare the success rates of LVTG, GelSlim, and DIGIT on tasks including grasping, insertion and removal, as summarized in Table~\ref{tab:sensor-c}. Due to its larger contact area and more distributed force application, LVTG achieves a higher success rate of 92$\%$ when grasping larger and heavier objects such as wine bottles, reducing the likelihood of slippage compared to GelSlim and DIGIT. In plate-grasping tasks, LVTG's wedge-shaped design allows it to slide more easily under the plate edge, resulting in a success rate of 89$\%$. In contrast, the cubic shapes of GelSlim and DIGIT make it more difficult to access the underside of the plate and maintain stable contact, leading to success rates of 81$\%$ and 73$\%$, respectively. For USB insertion and removal, where the small size of the connector enables all three sensors to grasp the plug head adequately, their performance is similar, with success rates of 76$\%$, 75$\%$, and 73$\%$, respectively. These results suggest that LVTG offers improved performance on certain everyday grasping tasks, particularly those involving object geometry or weight that challenge conventional sensor form factors.

\begin{table}[!t]
\centering
\caption{Success rates ($\%$) of different vision-based tactile sensors in teleoperated grasp stability and reliability tests}
\resizebox{\linewidth}{!}{ 
\begin{tabular}{ccccc}
\toprule
\begin{tabular}[c]{@{}c@{}}Vision-based \\ Tactile Sensors\end{tabular} & \begin{tabular}[c]{@{}c@{}}Grasping\\ Wine Bottle\end{tabular} & \begin{tabular}[c]{@{}c@{}}Grasping\\ Plate\end{tabular} & \begin{tabular}[c]{@{}c@{}}USB Insertion\\ and Removal\end{tabular} & \begin{tabular}[c]{@{}c@{}}Average \\ Scores\end{tabular} \\ \hline
GelSlim                                                                 & 85                                                             & 81                                                       & 76                                                                  & 81                                                      \\
DIGIT                                                                   & 80                                                             & 73                                                       & 75                                                                  & 76                                                      \\
LVTG                                                                    & 92                                                             & 89                                                       & 73                                                                  & 85                                                      \\ 
\bottomrule
\end{tabular}
}
\label{tab:sensor-c}
\end{table}

\subsection{Impact of Tactile Sensing on Policy Learning}

We further investigate whether the visuo-tactile feedback provided by LVTG enhances robot learning in contact-rich manipulation tasks. ACT is used as the baseline imitation learning framework. Following the protocol in \cite{george2025vital}, we compare three settings: (i) ACT with only visual RGB input; (ii) ACT with visual input and LVTG tactile images but without contrastive pretraining; and (iii) ACT with visual input, LVTG tactile images, and the proposed CLIP-style visuo-tactile pretraining. The policies are evaluated on five representative tasks, whose specifications are defined as follows. \textbf{Plate Wiping} requires the robot to grasp a cleaning sponge, establish stable contact with the plate surface, and remove visible stains without losing the sponge or pushing the plate out of the workspace. A trial is considered successful if the target stain region is visibly cleaned and the plate remains stable after the wiping motion. \textbf{Fragile Cup Manipulation} requires the robot to grasp a small plastic cup and place it onto a tray. Success is recorded only when the cup is stably placed within the tray without visible deformation, collapse, or dropping, thereby evaluating delicate object handling under limited allowable contact force. \textbf{Pouring Drinks} requires the robot to grasp a bottle, tilt it toward a receiving container, and complete the pouring motion while maintaining a stable grasp as the liquid distribution changes. A trial is successful if the drink is poured into the target container without bottle slip, excessive spillage, or loss of grasp. \textbf{USB Plugging} requires the robot to insert a pre-grasped USB plug into a specified socket. A trial is counted as successful if the plug is inserted to a sufficient depth and remains securely connected after release; tactile feedback is expected to assist fine alignment during contact. \textbf{Potato Chip Grasping} requires the robot to grasp a fragile chip and transfer it to a plate without breakage. Success is defined as placing the chip intact on the target plate without visible fracture or dropping, which tests whether the policy can maintain secure contact while avoiding excessive gripping force. Each policy is evaluated over 10 trials per task.

Implementation and training data collection are conducted on the COBOT Magic Robot platform~\cite{mu2024robotwin} equipped with LVTG as the end-effector visuo-tactile sensing module. The robot is connected to a workstation running Ubuntu 20.04 and ROS Noetic, and all representation learning, policy training, and policy inference experiments are accelerated using an NVIDIA RTX 4090 GPU. External RGB images and LVTG tactile images are published through ROS topics at 30 Hz, providing temporally synchronised visual and tactile observations for paired representation learning and downstream imitation learning. During policy evaluation, inference is performed on the GPU with an average latency below 20 ms per control cycle.
\begin{table}[!t]
\centering
\caption{Training hyperparameters used for visuo-tactile pretraining and ACT policy learning}
\resizebox{\linewidth}{!}{ 
\begin{tabular}{lll}
\toprule
Stage & Hyperparameter & Value \\
\midrule
\multirow{7}{*}{\begin{tabular}[c]{@{}l@{}}Contrastive\\ pretraining\end{tabular}} & Visual backbone & ResNet-18, frozen \\
 & Tactile backbone & ResNet-18, finetuned \\
 & Projection dimension & 256 \\
 & Optimiser & AdamW \\
 & Learning rate & $1\times10^{-4}$ \\
 & Batch size & 128 \\
 & Negative samples & In-batch unmatched pairs \\
\midrule
\multirow{10}{*}{\begin{tabular}[c]{@{}l@{}}ACT policy\\ learning\end{tabular}} & Visual input resolution & $640\times480$ \\
 & Tactile input resolution & $640\times480$ \\
 & Action chunk horizon $H$ & 48 \\
 & Hidden dimension & 512 \\
 & Visual backbone & ResNet-18 \\
 & Tactile backbone & Pretrained ResNet-18 \\
 & Learning rate & $4\times10^{-5}$ \\
 & Weight decay & $1\times10^{-4}$ \\
 & Batch size & 64 \\
 & KL weight $\beta$ & 10 \\
\bottomrule
\end{tabular}
}
\label{tab:training_details}
\end{table}

For the learning experiments, 1,000 teleoperated trajectories are collected for each task, resulting in 5,000 trajectories in total across the five tasks. Each trajectory contains time-aligned RGB frames, tactile frames, robot states, and operator commands. The same trajectory pool is used in two ways: timestamp-matched RGB--tactile frames are used to construct paired samples for contrastive representation pretraining, while the corresponding state--action sequences are used as demonstrations for ACT policy learning. All compared policy variants are trained with the same task split and demonstration set, so performance differences can be attributed to the input modality and the proposed visuo-tactile pretraining strategy rather than to differences in data volume. Both visual and tactile image streams are recorded at $640\times480$ resolution, balancing storage cost, training efficiency, and preservation of contact-induced tactile details.
\begin{table*}[!ht]
\centering
\caption{Detailed policy evaluation results on the COBOT Magic robot platform. Completion rate denotes the fraction of successful trials; task duration is the average execution time; slip count and damage count record visible object slips and object damage events; efficiency score is computed as completion rate divided by task duration.}
\resizebox{\textwidth}{!}{ 
\begin{tabular}{llccccc}
\toprule
Task & Method & \begin{tabular}[c]{@{}c@{}}Completion\\ Rate\end{tabular} & \begin{tabular}[c]{@{}c@{}}Task\\ Duration (s)\end{tabular} & \begin{tabular}[c]{@{}c@{}}Slip\\ Count\end{tabular} & \begin{tabular}[c]{@{}c@{}}Damage\\ Count\end{tabular} & \begin{tabular}[c]{@{}c@{}}Efficiency\\ Score\end{tabular} \\ \midrule
\multirow{3}{*}{Plate Wiping}
 & ACT (Vision-only) & 0.30 & 18.60 & 5 & 0 & 0.0161 \\
 & Ours (+Tactile, no pretraining) & 0.40 & 16.20 & 3 & 0 & 0.0247 \\
 & Ours (+Pretraining) & \textbf{0.60} & \textbf{14.50} & \textbf{1} & 0 & \textbf{0.0414} \\ \midrule
\multirow{3}{*}{Fragile Cup}
 & ACT (Vision-only) & 0.40 & 13.80 & 3 & 4 & 0.0290 \\
 & Ours (+Tactile, no pretraining) & 0.50 & 11.40 & 2 & 2 & 0.0439 \\
 & Ours (+Pretraining) & \textbf{0.60} & \textbf{10.20} & \textbf{1} & \textbf{1} & \textbf{0.0588} \\ \midrule
\multirow{3}{*}{Pouring Drinks}
 & ACT (Vision-only) & 0.30 & 16.50 & 4 & 0 & 0.0182 \\
 & Ours (+Tactile, no pretraining) & 0.40 & 14.30 & 2 & 0 & 0.0280 \\
 & Ours (+Pretraining) & \textbf{0.60} & \textbf{12.70} & \textbf{1} & 0 & \textbf{0.0472} \\ \midrule
\multirow{3}{*}{USB Plugging}
 & ACT (Vision-only) & 0.20 & 20.80 & 6 & 0 & 0.0096 \\
 & Ours (+Tactile, no pretraining) & 0.30 & 18.60 & 4 & 0 & 0.0161 \\
 & Ours (+Pretraining) & \textbf{0.40} & \textbf{17.10} & \textbf{3} & 0 & \textbf{0.0234} \\ \midrule
\multirow{3}{*}{Potato Chip}
 & ACT (Vision-only) & 0.30 & 15.70 & 3 & 5 & 0.0191 \\
 & Ours (+Tactile, no pretraining) & 0.50 & 13.50 & 2 & 3 & 0.0370 \\
 & Ours (+Pretraining) & \textbf{0.50} & \textbf{12.20} & \textbf{1} & \textbf{2} & \textbf{0.0410} \\ \bottomrule
\end{tabular}}
\label{tab:sr}
\end{table*}

The training pipeline is divided into representation pretraining and policy learning, as summarised in Table~\ref{tab:training_details}. This separation is important because the proposed method aims to test whether cross-modal visuo-tactile alignment improves downstream manipulation, rather than only increasing the capacity of the policy network. In the first stage, time-synchronised external RGB frames and calibrated tactile images are used to train the visual--tactile embedding space. The visual encoder is frozen and acts as a stable reference branch, while the tactile encoder and projection heads are updated to map contact-induced tactile images into the same latent space. The contrastive objective uses matched visual--tactile frames from the same timestamp as positive pairs and unmatched samples within the mini-batch as negatives.

The pretraining parameters in Table~\ref{tab:training_details} are selected to keep the alignment stage compact and reproducible. ResNet-18 is used for both visual and tactile backbones to match the moderate size of the collected dataset and to avoid introducing a much larger vision model as a confounding factor. Freezing the visual backbone reduces representation drift during contrastive learning, while finetuning the tactile backbone allows the network to adapt to the image statistics of the LVTG sensor. The 256-dimensional projection head provides a compact shared embedding for visual and tactile observations. A batch size of 128 supplies sufficient in-batch negative pairs for the CLIP-style loss without using a memory bank or external data.

For policy learning, the ACT stage uses the same image resolution, transformer hidden dimension, action chunk horizon, and optimisation settings across all compared variants. The action chunk horizon is set to $H=48$, so the policy predicts a short sequence of future joint commands rather than a single-step action. This setting is suitable for contact-rich tasks because tactile information often affects several subsequent control steps after contact is established. The pretrained tactile backbone is used only in the full model, whereas the no-pretraining tactile baseline uses the same policy architecture without the contrastive alignment stage.

% The ablation study isolates the contribution of tactile sensing and pretraining. The vision-only baseline tests whether external RGB observations are sufficient for the contact-rich tasks. The tactile variant tests the effect of directly adding LVTG tactile images to ACT. The pretrained variant evaluates whether the cross-modal representation in Eq.~\eqref{eq:contrastive_loss} provides more useful contact features for downstream control.

For negative sample construction, we avoid reliance on external datasets, memory banks, or complex sampling schemes and adopt the standard in-batch negative sampling strategy used in CLIP. Mild augmentations, including scaling, random cropping, and brightness perturbation, are applied to increase positive-pair diversity and improve robustness to local deformation, viewpoint changes, and sensor noise. Flipping operations are avoided because they may invert task-relevant spatial semantics. During contrastive learning, the paired visual-tactile sample at the same timestamp is defined as the positive sample, and all unmatched cross-modal pairs within the mini-batch serve as in-batch negative samples. This formulation encourages the model to learn shared contact-state-aware representations rather than overfitting to low-level visual appearance cues. Finally, to assess the effectiveness of the pretraining strategy, we compare the non-pretrained tactile baseline with the contrastively pretrained variant under the same ACT policy-learning protocol.

The experimental results are summarised in Table~\ref{tab:sr}. Following the detailed reporting style used in contact-rich manipulation studies, we report task completion rate, average task duration, slip count, damage count, and an efficiency score defined as completion rate divided by task duration. These metrics evaluate not only whether a policy completes the task, but also whether it does so efficiently and without unstable contact or object damage.

The vision-only ACT baseline achieves an average completion rate of 0.30 across the five tasks, with an average task duration of 17.08 s. It also produces 21 slip events and 9 damage events in total, showing that external RGB observations alone are insufficient for reliable contact-rich manipulation. Directly adding LVTG tactile observations increases the average completion rate to 0.42 and reduces the average duration to 14.80 s. The total slip and damage counts also decrease to 13 and 5, respectively, indicating that tactile feedback helps the policy maintain more stable contact and avoid excessive force. With CLIP-style visuo-tactile pretraining, the average completion rate further increases to 0.54 and the average duration decreases to 13.34 s. The pretrained policy also records the fewest slip events 7 and damage events 3, giving the best average efficiency score among the three policy variants.

The task-level results further clarify where tactile representation learning is most useful. In Plate Wiping and Pouring Drinks, the pretrained policy improves the completion rate to 0.60 and substantially reduces slip events, suggesting better regulation of broad surface contact and grasp stability during dynamic motion. In Fragile Cup Manipulation and Potato Chip Grasping, the reduction in damage count shows that tactile cues help the policy avoid overly aggressive contact when handling deformable or brittle objects. USB Plugging remains the most difficult task: although pretraining improves the completion rate from 0.20 to 0.40 and reduces slip count from 6 to 3, the absolute completion rate is still lower than in the other tasks because fine insertion requires precise pose alignment after contact. Overall, Table~\ref{tab:sr} shows that tactile observations improve manipulation reliability, while contrastive pretraining makes those observations more useful for downstream ACT policy learning.

While the proposed system achieves strong overall performance, we observe several representative failure modes that reveal its current limitations. First, in low-friction or highly slippery conditions (e.g., plastic film or polished surfaces), the contact signals become weak and unstable, leading to inaccurate estimation of contact quality and occasional object slip during manipulation. Second, for objects with challenging geometries (e.g., thin, highly curved, or partially transparent structures), the effective contact area is limited or uneven, producing ambiguous tactile patterns and reducing the reliability of contact localization, as observed in fine alignment tasks such as USB insertion. In addition, large-force contacts may cause non-linear deformation or local saturation of the elastomer, which reduces sensitivity to fine-grained variations. Finally, in scenarios with severe visual occlusion combined with weak or intermittent contact, the multimodal representation becomes insufficient to reliably infer interaction states.
\begin{table*}[!th]
\centering
\caption{Policy task completion rates ($\%$) for contact-rich tasks on the Lebai LM3 robot platform}
\resizebox{\textwidth}{!}{ 
\begin{tabular}{ccccccc}
\toprule
Method                         & Plate Wiping & \begin{tabular}[c]{@{}c@{}}Fragile Cup \\ Manipulation\end{tabular} & Pouring Drinks & USB Plugging & \begin{tabular}[c]{@{}c@{}}Potato Chip \\ Grasping\end{tabular} & \begin{tabular}[c]{@{}c@{}}Average \\ Completion Rate\end{tabular} \\ \hline
ACT (Vision-only)               & 0.31           & 0.36                                                                  & 0.28              & 0.27           & 0.32                                                             & 0.31                                                        \\
Ours (+Tactile, no pretraining) & 0.46          & 0.53                                                                  & 0.41             & 0.35          & 0.42                                                            & 0.43                                                       \\
Ours (+Pretraining)             & \textbf{0.58}           & \textbf{0.63}                                                                  & \textbf{0.56}             & \textbf{0.42}           & \textbf{0.55}                                                              & \textbf{0.55}                                                        \\ \bottomrule
\end{tabular}}
\label{tab:lebai}
\end{table*} 

To assess platform compatibility, we conduct preliminary experiments on the Lebai LM3 robot in addition to the COBOT Magic platform. The LM3 is a widely used platform for industrial automation, with a six-axis articulated arm and advanced control capabilities \cite{lebai}. We repeat the same manipulation tasks after integrating LVTG with an LM3-compatible adaptor. Because these experiments are intended as a transferability check rather than the main benchmark, we report only task completion rates. As shown in Table~\ref{tab:lebai}, the relative ranking of the three policy variants is consistent with that observed on the COBOT Magic platform. These results suggest that the modular LVTG design can be transferred across robot platforms, although broader validation on more embodiments remains future work.

\section{Conclusion} \label{Sec: conclusion}
In this paper, we presented a cross-modal representation learning pipeline for contact-rich manipulation. The tactile platform provides paired observations for training, and the learned embeddings improve ACT policy performance over vision-only and non-pretrained tactile baselines. The current results support the algorithmic value of cross-modal pretraining, while the hardware serves as a reproducible sensing front-end for the learning study. The current design uses a flat sensing surface on a parallel-jaw gripper, which simplifies manufacturing and calibration but limits conformity to complex geometries and reduces contact richness. Extending to curved or non-planar tactile surfaces introduces challenges in optical calibration and deformation modeling. Future work will explore multi-view designs and modular curved sensing units for more dexterous manipulation.

\section*{Funding}
 This work is supported by the Science and Technology Development Fund, Macau, SAR, under Grant 0005/2023/EIB1.

\section*{Data Availability}
The project is available at \url{https://github.com/huazai665/LVTG}. 

\end{document}